# Moving Beyond the Turing Test with the Allen AI Science Challenge

*Carissa Schoenick, Peter Clark, Oyvind Tafjord, Peter Turney, Oren Etzioni*
*April 2016*

## Introduction

The field of Artificial Intelligence has made great strides forward recently, for example AlphaGo's recent victory against the world champion Lee Sedol in the game of Go, leading to great optimism about the field. But are we really moving towards smarter machines, or are these successes restricted to certain classes of problems, leaving other challenges untouched? In 2016, the Allen Institute for Artificial Intelligence (AI2) ran the Allen AI Science Challenge, a competition to test machines on an ostensibly difficult task, namely answering 8th Grade science questions. Our motivations were to encourage the field to set its sights broader and higher by exploring a problem that appears to require modeling, reasoning, language understanding, and commonsense knowledge, to probe the state of the art on this task, and sow the seeds for possible future breakthroughs. The challenge received a strong response, with 780 teams from all over the world participating. What were the results? This article describes the competition and the interesting outcomes of the challenge.

## Motivation

Challenge problems play an important role in motivating and driving progress in a field. For a field striving to endow machines with intelligent behavior, e.g., language understanding and reasoning, challenge problems that test such skills are essential.

In 1950, Alan Turing proposed the now well-known Turing Test as a possible test of machine intelligence: if a system can exhibit conversational behavior that is indistinguishable from that of a human during a conversation, that system could be considered intelligent [1]. As the field of AI has grown, this Test has become less meaningful as a challenge task for several reasons. First, in its details, it is not well-defined, e.g., who is the person giving the test? A computer scientist would likely know good distinguishing questions to ask, while a random member of the population may not. What constraints are there on the interaction? What guidelines are provided to the judges? Second, recent Turing Test competitions have shown that, in certain formulations, the Turing Test is gameable - people can be fooled by systems that simply retrieve sentences, and make no claim of being intelligent [2,3]. As *The New York Times*'s John Markoff puts it, the Turing Test is more a test of human gullibility than machine intelligence. Finally, the test, as originally conceived, is pass/fail rather than scored, thus providing no measure of progress towards a goal, something essential for a challenge problem[1][2].

---

[1] Indeed, Turing himself did not conceive of the Turing Test as a challenge problem to drive the field forward, but rather as a thought experiment about a useful alternative to the question of "Can machines think?".
[2] Although one can imagine metrics that quantify performance on the Turing Test, the imprecision in the task definition and human variability makes it hard to define metrics that are reliably reproducible.

Nowadays, machine intelligence is viewed less as a binary pass/fail attribute, and more as a diverse collection of capabilities associated with intelligent behavior. Rather than a single test, cognitive scientist Gary Marcus of NYU and others recently proposed the notion of series of tests, a Turing Olympics of sorts, that could assess the full gamut of AI from robotics to NLP [4][5].

Our goal with the Allen AI Science Challenge was to operationalize one such test, namely answering science exam questions. Clearly the Science Challenge is not a full test of machine intelligence. However, it does explore several capabilities strongly associated with intelligence - capabilities that our machines need if they are to reliably perform the smart activities we desire of them in the future - including language understanding, reasoning, and use of commonsense knowledge. Doing well on the challenge appears to require significant advances in AI technology, making it a potentially powerful vehicle for advancing the field. In addition, from a practical point of view, exams are accessible, measurable, understandable, and compelling.

One of the most interesting and appealing aspects of science exams is their graduated and multifaceted nature: different questions explore different types of knowledge, and they vary substantially in difficulty (especially for a computer). There are questions that can be easily addressed with a simple fact lookup, like this one:

> How many chromosomes does the human body cell contain?
> (A) 23
> (B) 32
> (C) 46
> (D) 64

And then there are questions requiring extensive understanding of the world, such as this example:

> City administrators can encourage energy conservation by
> (A) lowering parking fees
> (B) building larger parking lots
> (C) decreasing the cost of gasoline
> (D) lowering the cost of bus and subway fares

This question requires the knowledge that certain activities and incentives result in human behaviors, which in turn result in more or less energy being consumed. Understanding this question also requires recognizing that "energy" in this context refers to resource consumption for the purposes of transportation (as opposed to other forms of energy one might find in a science exam, like electrical, kinetic/potential, etc.).

# AI vs 8th Grade: The Allen AI Science Challenge

To put this approach to the test, AI2 designed and hosted "The Allen AI Science Challenge," a four-month long competition in partnership with Kaggle.com that concluded in February of 2016 [7]. Researchers worldwide were invited to build AI software that could answer standard 8th grade multiple choice science questions. The competition aimed to assess the state of the art in AI systems utilizing natural language understanding and knowledge-based reasoning—how accurately the participants' models could answer the exam questions would serve as an indicator of how far the field has come in these areas.

Competition Overview

**Timeline and Participants**
The competition lasted four months from October 7th, 2015 through February 13th, 2016. A total of 780 teams participated during the model building phase, and 170 teams made a final model submission. Participants were required to make the code for their model available to AI2 at the close of the competition to validate model performance and to confirm the models followed contest rules. At the conclusion of the competition, the winners were also expected to make their code open source. The three teams that achieved the highest scores on the challenge's test set received prizes of $50,000, $20,000, and $10,000 respectively.

**Data**
A total of 5,083 8$^{th}$ grade multiple choice science questions were licensed from providing partners for the purposes of the competition. All questions were standard multiple choice format with four answer options, as the examples provided above. From this collection of questions, participants were provided with a set of 2,500 training questions with which to train their models. A validation set of 8,132 questions was used during the course of the competition for confirming model performance. Only 800 of the validation questions were legitimate; the rest were artificially generated to disguise the real questions in order to prevent cheating via manual question answering or unfair advantage of additional training examples. A week before the end of the competition, the final test set of 21,298 questions (which also included the validation set) were provided to participants to use to produce a final score for their model (of these, 2,583 questions were legitimate). The data for the competition was licensed from private assessment content providers who did not wish to allow the use of their data beyond the constraints of the competition, however AI2 has made some subsets of these questions available on their website [8].

**Baselines and Scores**
As these questions are all 4-way multiple choice, a standard baseline score using random guessing is 25%. AI2 also generated a baseline score using a Lucene search over the Wikipedia corpus, which produced scores of 40.2% on the training set and 40.7% on the final test set. The final outcome of the competition was quite close, with the top three teams achieving scores with a spread of only 1.05%. The highest score was 59.31%.

First Place

Top prize went to Chaim Linhart of Israel (Kaggle username Cardal). His model achieved a final score of 59.31% on the test question set using a combination of 15 gradient boosting models, each of which used a different subset of features. Unlike the other winners' models, Chaim's model predicts the correctness of each answer option individually. There were two general categories of features used to make these predictions; the first category was made up of information retrieval (IR) based features, applied by searching over corpora he compiled from various sources such as study guide or quiz building websites, open source textbooks, and Wikipedia. His searches used various weightings and stemmings to optimize performance. The other flavor of feature used in his ensemble of 15 models was based on properties of the questions themselves, such as the length of the question and answer, the form of the answers (e.g., characteristics like numeric answer options, answers that contained referential clauses like "none of the above" as an option), and the relationships between answer options.

Chaim explained that he used several smaller gradient boosting models instead of one big model in order to maximize diversity. One big model tends to ignore some important features because it requires a very large training set to require it to pay attention to all of the potentially useful features present--using several small models requires that the learning algorithm use features that it would otherwise ignore, given the more limited training data available in this competition.

The IR-based features alone could achieve scores as high as 55% by Chaim's estimation. His question-form features fill in some remaining gaps to bring the system up to about 60% correct. The 15 models were combined by a simple weighted average to yield the final score for each choice. Chaim credited careful corpus selection as one of the primary elements driving the success of his model.

## Second Place

The second place team with a score of 58.34% was a group of people from a social media analytics company based in Luxembourg called Talkwalker, led by Benedikt Wilbertz (Kaggle username poweredByTalkwalker).

Benedikt's team built a relatively large corpus as compared to other winning models, which used 180GB of disk space after indexing with Lucene. They utilized several feature types, including IR-based features using their large corpus, vector-based features (scoring question-answer similarity by comparing vectors from word2vec and Glove), pointwise mutual information (PMI) features (measured between the question and target answer, calculated on their large corpus), and string hashing features in which term-definition pairs were hashed and then a supervised learner was trained to classify pairs as correct or incorrect. A final model uses these various features to learn pairwise ranking between the answer options using the XGBoost gradient boosting library.

The use of string hashing features by the poweredByTalkwalker team is unique; this methodology was not tried by either of the other two competition winners, nor is it used in AI2's Project Aristo. The team used a corpus of terms and definitions obtained from an educational flashcard building site, and then created negative examples by mixing terms with random definitions. A supervised classifier was trained on these incorrect pairs, and then the output was used to generate features for input to XGBoost.

## Third Place

The third place winner was Alejandro Mosquera from the UK (Kaggle username Alejandro Mosquera), with a score of 58.26%. Alejandro approached the challenge as a three-way classification problem for each pair of answer options. The choices A, B, C, and D were transformed to all twelve pairs (A,B), (A,C), ..., (D,C), which were labeled with three classes, the *left* pair element is correct, the *right* is correct, or *neither* is correct. The pairs were then classified using logistic regression. This three-way classification is easier for supervised learning algorithms than the more natural two-way (*correct* versus *incorrect*) classification with four choices, because the two-way classification requires an absolute decision about a choice, whereas the three-way classification requires only a relative ranking of the choices. Alejandro made use of three types of features: IR-based features based on scores from Elastic Search using Lucene over a corpus, vector-based features that measured question-answer similarity by comparing vectors from word2vec, and question-form features that considered things such as the structure of a question, the length of the question and the answer choices. Alejandro also noted that careful corpus selection was crucial to his model's success.

## Competition Lessons

In the end, each of the winning models found the most benefit in information retrieval based methods. This is indicative of the state of AI technology in this area of research; we can't ace an 8th grade science exam because we do not currently have AI systems capable of going beyond the surface text to a deeper understanding of the meaning underlying each question, and then successfully using reasoning to find the appropriate answer. All three winners expressed that it was clear that applying a deeper, semantic level of reasoning with scientific knowledge to the questions and answers would be the key to achieving scores of 80% and beyond, and to demonstrating what might be considered true artificial intelligence.

A few other example questions from the competition that each of the top three models got wrong highlight the more interesting, complex nuances of language and chains of reasoning an AI system will need handle in order to answer these questions correctly, and for which IR methods aren't sufficient:

> What do earthquakes tell scientists about the history of the planet?
> (A) Earth's climate is constantly changing.
> (B) The continents of Earth are continually moving.
> (C) Dinosaurs became extinct about 65 million years ago.
> (D) The oceans are much deeper today than millions years ago.

This question digs into the causes behind earthquakes and the larger geographic phenomena of plate tectonics, and cannot be easily solved by looking up a single fact. Additionally, other true facts appear in the answer options ("*Dinosaurs became extinct about 65 million years ago.*"), but must be intentionally identified and discounted as being incorrect in the context of the question.

> Which statement correctly describes a relationship between the distance from Earth and a characteristic of a star?
> (A) As the distance from Earth to the star decreases, its size increases.

(B) As the distance from Earth to the star increases, its size decreases.
(C) As the distance from Earth to the star decreases, its apparent brightness increases.
(D) As the distance from Earth to the star increases, its apparent brightness increases.

This question requires general common-sense type knowledge of the physics of distance and perception, as well as the semantic ability to relate one statement to another within each answer option to find the right directional relationship.

## Other Attempts

While there are numerous question-answering systems that have emerged from the AI community, none address the challenges of scientific and commonsense reasoning exhibited by the example questions above. Question-answering systems developed for the MUC (message understanding) conferences [9] and TREC (text retrieval) conferences [10] focused on retrieving answers from text, the former from newswire articles and the latter from various large corpora such as the Web, microblogs, and clinical data. More recent work has focused on answer retrieval from structured data, e.g., "In which city was Bill Clinton born?" from FreeBase [11,12,13]. These systems rely on the information being stated explicitly in the underlying data, however, and are unable to perform the reasoning steps that would be required to conclude this information from indirect supporting evidence.

There are a few systems that attempt some form of reasoning: Wolfram Alpha [14] answers mathematical questions, providing they are stated either as equations or with relatively simple English; Evi [15] is able to combine facts together to answer simple questions (e.g., Who is older, Barack or Michelle Obama?); and START [16] will similarly answer simple inference questions using Web-based databases (e.g., What South-American country has the largest population?). However, none of these systems attempt the level of complex question processing and reasoning that will be required to successfully answer many of the science questions in the Allen AI Challenge.

## Looking Forward

As the 2015 Allen AI Science Challenge clearly demonstrates, achieving a high score on a science exam is going to require a system that can do more than merely sophisticated information retrieval. Project Aristo at AI2 is intently focused on this problem of successfully demonstrating artificial intelligence using standardized science exams, developing an assortment of approaches to address the challenge. AI2 plans to release additional data sets and software for the wider AI research community to utilize in this effort [8].

## References

[1] Alan M. Turing. Computing machinery and intelligence. Mind, LIX(236):433–460, October 1950.
[2] BBC. Computer AI passes Turing test in 'world first' Turing Test. BBC News. 9 June 2014. http://www.bbc.com/news/technology-27762088

[3] Aron, J. Software tricks people into thinking it is human. New Scientist (Issue 2829), Sept 2011.
[4] Marcus, G., Rossi, F., Veloso, M. (Eds), Beyond the Turing Test (AI Magazine Special Edition), AI Magazine, 37 (1), Spring 2016.

[5] Turk, V. The Plan to Replace the Turing Test with a 'Turing Olympics'. Motherboard. 28 January 2015. http://motherboard.vice.com/read/the-plan-to-replace-the-turing-test-with-a-turing-olympics
[6] Clark, P., Etzioni, O. My Computer is an Honor Student - But how Intelligent is it? Standardized Tests as a Measure of AI. In AI Magazine 37 (1), Spring 2016.
[7] "The Allen AI Science Challenge." Kaggle, 13 Feb. 2016, https://www.kaggle.com/c/the-allen-ai-science-challenge.
[8] "Data - Allen Institute for Artificial Intelligence." Allen Institute for Artificial Intelligence, http://allenai.org/data. Accessed 20 Oct. 2016.
[9] Grishman, R., Sundheim, B. Message Understanding Conference-6: A Brief History. In COLING (Vol. 96, pp. 466-471), 1996.
[10] Voorhees, E., Ellis, A. (Eds)Proc. 24th Text REtrieval Conference (TREC 2015), Publication SP 500-319, NIST (http://trec.nist.gov/ ), 2015.
[11] Yao, X., Van Durme, B. Information Extraction over Structured Data: Question Answering with Freebase. In ACL (1) (pp. 956-966), 2014.
[12] Berant, J., Chou, A., Frostig, R., Liang, P. Semantic Parsing on Freebase from Question-Answer Pairs. In EMNLP (Vol. 2, No. 5, p. 6), 2013.
[13] Fader, A., Zettlemoyer, L., & Etzioni, O. Open question answering over curated and extracted knowledge bases. In Proc 20th ACM SIGKDD Int Conf on Knowledge Discovery and Data Mining (pp. 1156-1165). ACM, 2014.
[14] Wolfram, S. Making the World's Data Computable. Proc. Wolfram Data Summit, 2010. (http://blog.stephenwolfram.com/2010/09/making-the-worlds-data-computable/ )
[15] Simmons, J. True Knowledge: The Natural Language Question Answering Wikipedia for Facts. In: Semantic Focus, Feb 2008.
[16] Katz, B., Borchardt, G., Felshin, S. Natural Language Annotations for Question Answering. Proc 19th Int FLAIRS Conference (FLAIRS 2006), 2006. (http://start.csail.mit.edu)
[17] Sachan, M., Dubey, A., Xing, E. Science Question Answering using Instructional Materials. arXiv preprint at arXiv:1602.04375  http://arxiv.org/pdf/1602.04375.pdf